\documentclass[runningheads]{llncs}
\usepackage{graphicx}
\usepackage{amsmath,amssymb} % define this before the line numbering.
\usepackage{color}

\usepackage{subfigure}

\DeclareMathOperator*{\argmax}{arg\,max}

\newcommand{\ranker}{Scorer}

\newcommand{\eg}{\emph{e.g.}} 
\newcommand{\ie}{\emph{i.e.}}

\newcommand{\etal}{\emph{et al}.}

\begin{document}
\pagestyle{headings}
\mainmatter

\def\ACCV20SubNumber{0568}  % Insert your submission number here

%===========================================================
\title{Addressing Class Imbalance in Scene Graph Parsing by Learning to Contrast and Score} % Replace with your title
\titlerunning{Scene Graph Parsing by Learning to Contrast and Score}
\author{He Huang\inst{1}\thanks{Part of this work was done while He Huang was an intern at Preferred Networks.} \and
Shunta Saito\inst{2} \and
Yuta Kikuchi\inst{2} \and
Eiichi Matsumoto\inst{2} \and
Wei Tang\inst{1} \and
Philip S. Yu\inst{1}
}
\authorrunning{H. Huang et al.}
\institute{University of Illinois at Chicago, Chicago, USA \\
\email{\{hehuang, tangw, psyu\}@uic.edu} \and
Preferred Networks Inc., Tokyo, Japan \\
\email{\{shunta, kikuchi, matsumoto\}@preferred.jp}\\
}

\maketitle

%===========================================================
\begin{abstract}
Scene graph parsing aims to detect objects in an image scene and recognize their relations. Recent approaches have achieved high \emph{average} scores on some popular benchmarks, but fail in detecting rare relations, as the highly long-tailed distribution of data biases the learning towards frequent labels. Motivated by the fact that detecting these rare relations can be critical in real-world applications, this paper introduces a novel integrated framework of classification and ranking to resolve the class imbalance problem in scene graph parsing. Specifically, we design a new Contrasting Cross-Entropy loss, which promotes the detection of rare relations by suppressing incorrect frequent ones. Furthermore, we propose a novel scoring module, termed as Scorer, which learns to rank the relations based on the image features and relation features to improve the recall of predictions. Our framework is simple and effective, and can be incorporated into current scene graph models. Experimental results show that the proposed approach improves the current state-of-the-art methods, with a clear advantage of detecting rare relations.
\end{abstract}

%===========================================================

\section{Introduction}
\begin{figure}[!t]
\centering
\subfigure[Distribution of the 50 most frequent predicates in VG~\cite{krishna2017visual_genome}.]{
\begin{minipage}[l]{0.8\linewidth}
\centering
\includegraphics[width=0.9\textwidth]{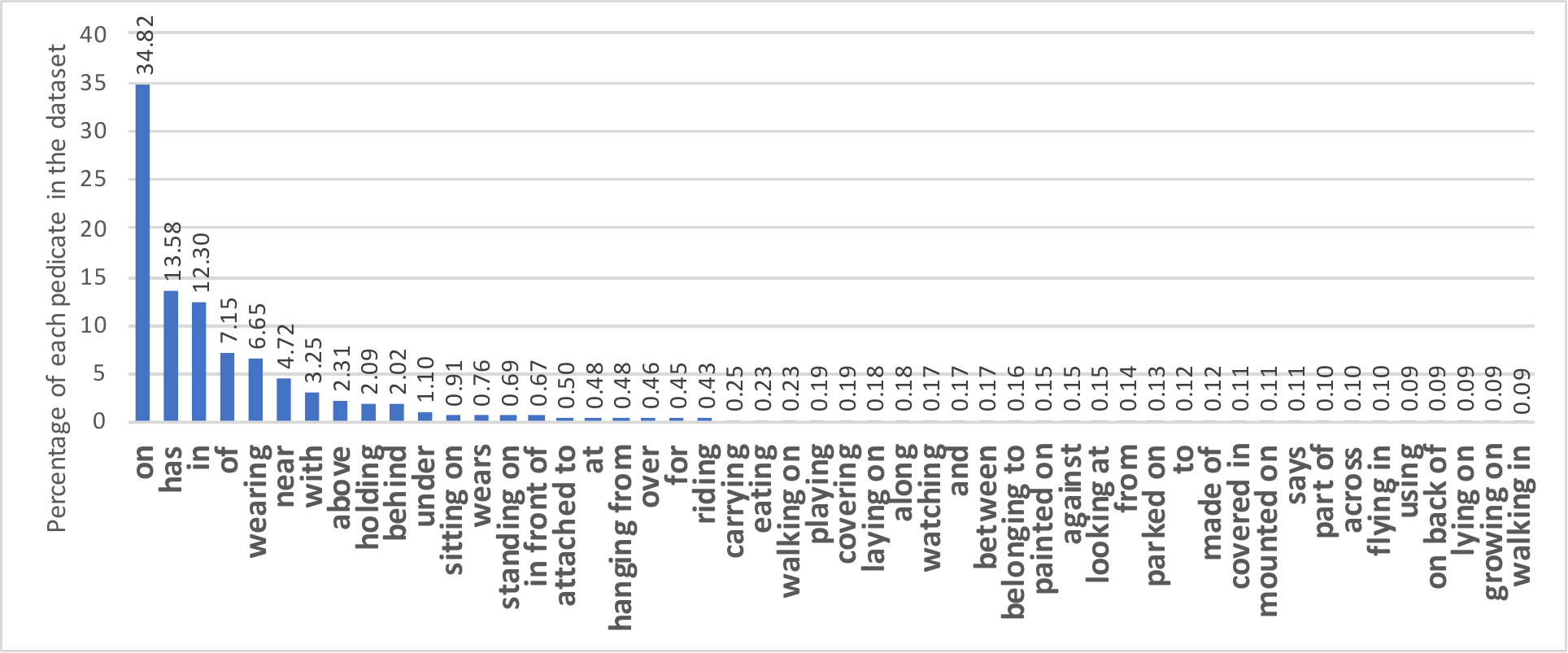}
\end{minipage}
\label{fig:stats}
}
\subfigure[Per-class Recall@100 of LinkNet~\cite{woo2018linknet} and MotifNet~\cite{zellers2018motif} evaluated on the predicate classification task.]{
\begin{minipage}[l]{0.8\linewidth}
\centering
\includegraphics[width=0.9\textwidth]{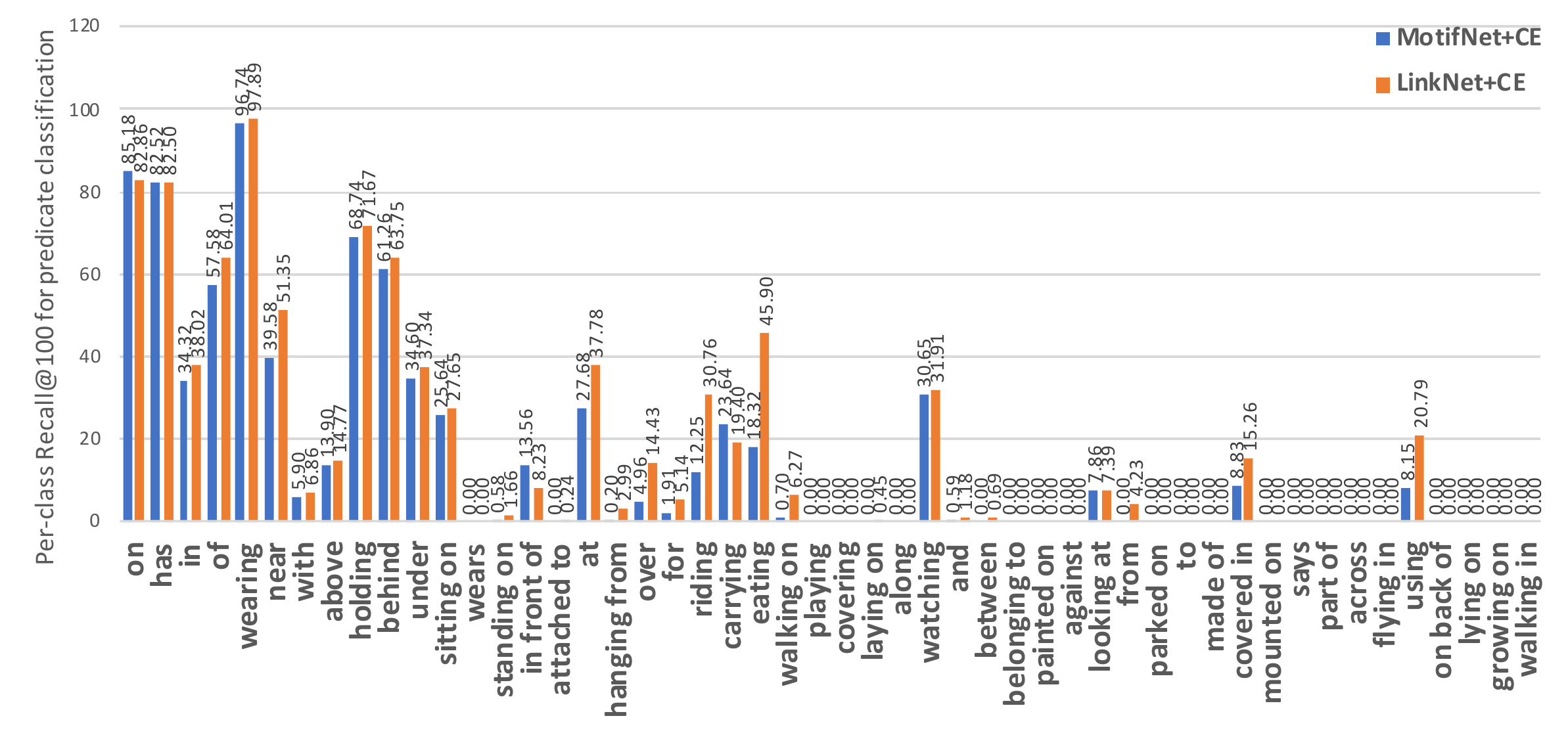}
\end{minipage}
\label{fig:backbone-ce}
}
\caption{Motivation of this work: (a) the predicate distribution is highly long-tailed,
and (b) two current state-of-the-art models~\cite{zellers2018motif,woo2018linknet} perform poorly on rare relations.}
\label{fig:vis}
\end{figure}
%Meanwhile, as in Figure~\ref{fig:compare-all}, our proposed CCE+\ranker\ framework can significantly improve the two models' performance on rare classes.
%One of the ultimate goals of computer vision is to teach machines how to understand the world using visual information such as images and videos. With advances in deep learning techniques~\cite{krizhevsky2012imagenet_conv,hochreiter1997lstm, kipf2016GCN} and large-scale datasets~\cite{deng2009imagenet}, machines are now able to accurately classify images from thousands of classes. Object detection~\cite{ren2015faster, redmon2016yolo, liu2016ssd}
%pushes the boundary of machines' capability a step forward by utilizing pretrained image classifiers and learning how to localize and classify different objects in given images. In this way, machines can understand the world not only from the image-level, but also from individual object's level. As an extension to object detection, scene graph detection~\cite{xu2017IMP, lu2016visual} requires machines to classify not only objects and but also the relationships between them, given the object proposals from pretrained object detectors.
As an extension to object detection~\cite{ren2015faster}, scene graph parsing  ~\cite{xu2017IMP,lu2016visual} aims to detect not only objects,
\eg, \emph{persons} and \emph{bikes}, in
an image scene but also recognize their relationships (also called \emph{predicates}), \eg, \emph{ride} and \emph{push}.
It is a fundamental tool for several applications such as image captioning~\cite{yang2019auto}, image retrieval~\cite{johnson2015retrieval} and image generation~\cite{johnson2018generation}.
Due to the combinatorially large space of valid relation triplets \texttt{<subject, predicate, object>} and the polysemy of a predicate in different contexts, the task of scene graph parsing is challenging.

%Most of existing scene graph models aim at learning better relation features for predicate classification, by designing more sophisticated networks~\cite{woo2018linknet,wang2019exploring} or incorporating more prior knowledge~\cite{zellers2018motif,chen2019KERN}. Although these methods generally work well, the improvement of recent methods on the predicate classification task in the past year is about $1\%$, with the highest being around $Recall@100=67.6\%\sim 68\%$~\cite{chen2019KERN, zhang2019graphical, woo2018linknet}. Such limited improvement seems to imply that scene graph detection has reached its bottleneck.
Recent scene graph parsing systems~\cite{xu2017IMP,zellers2018motif,woo2018linknet,chen2019KERN,yang2018graph,dai2017relational} are built on deep neural
networks due to their ability to learn robust feature representations for both images and relational contexts directly from data. Xu~\etal~\cite{xu2017IMP} use GRU~\cite{cho2014GRU} to approximate a conditional random field to jointly infer objects and predicates. Zellers~\etal~\cite{zellers2018motif} propose to utilize the statistical distribution of relation triplets as external prior knowledge. Yang~\etal~\cite{yang2018graph} propose to incorporate a GCN~\cite{kipf2016GCN} for predicting the predicates.

%By designing more sophisticated network architectures
%\cite{woo2018linknet,wang2019exploring} and including prior knowledge of the
%relationships \cite{zellers2018motif,chen2019KERN}, they have
%significantly pushed forward the state-of-the-art performance on this task.

%In order to find out the bottleneck, we first investigate the 50 most frequent predicates' distribution in the most widely used scene graph dataset Visual Genome (VG)~\cite{krishna2017visual_genome}. As is illustrated in Figure~\ref{fig:stats}, the distribution is highly long-tailed, with the top-5 predicates covering 74.5\% of all samples. Then for each predicate we calculate its recall at the top-100 predictions produced by two state-of-the-art methods LinkNet~\cite{woo2018linknet} and MotifNet~\cite{zellers2018motif} for the predicate classification task, as illustrated in  Figure~\ref{fig:backbone-ce}.

While these approaches can achieve high \emph{average} scores on some popular
benchmarks, their effectiveness is largely jeopardized in case of rare relationships.
For example, Figure \ref{fig:stats} and Figure \ref{fig:backbone-ce}
respectively show the distribution of predicates in the  Visual Genome dataset (VG) \cite{krishna2017visual_genome}
and the performances of two state-of-the-art methods, \ie,
LinkNet~\cite{woo2018linknet} and MotifNet~\cite{zellers2018motif}.
Both approaches perform quite well on frequent relations, \eg, \texttt{on} and \texttt{has}, 
but their performances degrade significantly on
less common predicates, \eg, \texttt{lying on} and \texttt{flying in}.
The recalls of several rare relations are even zero.
Actually, the average recalls of each class (\ie, \emph{macro-averaged Recall@100}) are only 15.3\% and 16.9\% for MotifNet~\cite{zellers2018motif}
and LinkNet~\cite{woo2018linknet} respectively.
By contrast, they can achieve around 67\% \emph{micro-averaged Recall@100}, which is calculated in a class-agnostic way.
In other words, a high micro-averaged recall is achieved 
at the cost of sacrificing the performance on rare classes.

This is because the distribution of the predicates is highly long-tailed, and
learning therefore tends to bias towards frequent relations so as to boost the overall
performance.
However, we argue that detecting those rare or \emph{abnormal} relations is critical in practice.
For example, consider a common predicate \texttt{on} and a rare one \texttt{lying on} in the VG dataset. 
Although these two terms are closely related, 
it is crucial for a surveillance camera to distinguish between \texttt{<person, on, street>} and \texttt{<person, lying on, street>}, since
the latter may imply an emergency that requires immediate reactions.

Another limitation of previous approaches is that they model the task as a pure classification problem, by training with two cross-entropy loss functions for \texttt{subjects/objects} and \texttt{predicates} respectively. 
This approach is problematic as the evaluation metric is recall@K, which not only depends on classification accuracy but also is sensitive to the ranks of predicted relation triplets.
While an object may either exist or not in an image, there  is always at least one relationship between any two existing objects, \eg, \texttt{left} or \texttt{right}. When people annotate the VG~\cite{krishna2017visual_genome} dataset, they only label the most important or nontrivial relations but ignore the others. 
As a result, the detection of an existing but trivial relation will be counted as a \emph{false positive} because it is unlabeled in the dataset.
To achieve a high recall for a given number of predictions, we need to not only classify the relations correctly but also rank the nontrivial relations higher than the trivial ones.

To overcome these challenging issues, this paper introduces a novel method which solves predicate classification and relation ranking in a unified framework.
We first introduce a loss function, termed as the Contrasting Cross-Entropy (CCE) loss, to handle the class imbalance problem.
This is achieved by simultaneously maximizing the predicted probability of the correct class and minimizing that of the hardest negative class.
By suppressing incorrect frequent relations, the CCE loss promotes the recall of rare predicates.
Furthermore, we propose a novel network module, called Scorer, to tackle the ranking problem.
%It produces a score of each predicted relation triplet by 
%exploiting context information of the object pair as well as the predicate distribution.
It scores a predicted relation triplet by comparing it with all other triplets as well as exploiting the global context information.
Ablation study indicates that (1) the CCE loss significantly improves the detection of rare relationships and (2) the Scorer network helps boost the recall by learning to rank. Experimental results show that the proposed CCE loss and Scorer network can improve the state-of-the-art methods on three tasks.
The contributions of this paper are:
\begin{itemize}
\itemsep0em
\iffalse
    \item We propose an novel integrated classification and ranking framework for scene graph detection.
    %\item We propose to 
    %formulate the scene graph detection problem as a combination of both classification and ranking problems.
    \item We propose a simple but effective Contrasting Cross-Entropy (CCE) loss to alleviate the class imbalance problem. It significantly improves the detection performance on rare classes.
    \item We propose a \ranker\ network to tackle the ranking problem. It can effectively utilize relation features generated by arbitrary scene graph models and the image features from pretrained image classifiers.
    \item We conduct extensive experiments and show that our combined CCE-\ranker\ framework with LinkNet~\cite{woo2018linknet} as the base scene graph model achieves new state-of-the-art results on three scene graph detection tasks. 
    \item In addition, our framework is light-weighted and can be easily applied to any existing scene graph models that are trained with cross-entropy.
\fi
\item We provide a brand-new perspective
on Scene Graph Parsing (SGP). To our knowledge, this
is the first attempt to formulate it as a joint task of classification and ranking.
\item We propose a simple but effective Contrasting Cross-Entropy (CCE) loss to alleviate the class imbalance problem. By contrasting for each relation the predicted probabilities of the correct label and the hardest incorrect one, it suppresses the incorrect frequent relations and significantly improves the detection performance on rare classes.
\item We introduce a novel \ranker\ network to tackle the ranking problem and improve the recall of baselines by a large margin.
It innovatively bridges the
point-wise and list-wise ranking approaches to take advantages
of both. Specifically, our \ranker\ exploits a self-attention layer
to compare one relation triplet with all the others in a listwise
fashion, from which learning to rank is achieved via a
point-wise cross-entropy loss. To our knowledge, this unification is unique and novel.
\item 
Our novel framework is general and light-weighted. It can be easily applied to any scene graph models trained with cross-entropy.
We conduct extensive experiments and show that our combined CCE-\ranker\ framework with LinkNet~\cite{woo2018linknet} as the base scene graph model achieves new state-of-the-art results on three scene graph detection tasks. 
\end{itemize}

\section{Related Work}
% The concept of scene graph is first applied in image retrieval~\cite{johnson2015retrieval}, where a Conditional Random Field (CRF) is utilized to model and pass information between object nodes in a scene graph to learn to ground them into given images. As the first model in scene graph generation, the IMP~\cite{xu2017IMP} uses GRUs to approximate CRF and proposes a message passing mechanism that updates the objects' and their edges' features iteratively. Besides, scene graph is also studied in other fields such as image generation~\cite{johnson2018generation}, image captioning~\cite{yao2018exploring} and few-shot learning~\cite{dornadula2019fewshot}.
\textbf{Scene Graph Parsing}. There have been several lines of research on scene graph parsing~\cite{xu2017IMP,herzig2018mapping,li2018factorizable} and visual relationship detection~\cite{lu2016visual,yu2017distillation} over the past few years. 
% Although they have different names, these two tasks are essentially the same, where both aim to detect the relationships between pairs of objects from given images. 
% Scene graph is also studied in other fields such as image generation~\cite{johnson2018generation}, image captioning~\cite{yao2018exploring, yang2019auto, gu2019unpaired} and few-shot learning~\cite{dornadula2019fewshot}. 
IMP~\cite{xu2017IMP} uses GRUs~\cite{cho2014GRU} to approximate a CRF and proposes an iterative message passing mechanism. 
% Lu \etal~\cite{lu2016visual} propose to use linguistic knowledge extracted from pretrained language models to predict relations based on the word embeddings of associated objects.
% Yu \etal~\cite{yu2017distillation} collect linguistic knowledge of $P(predicate|subject, object)$ from both training dataset as well as the Wikipedia website, and propose to use a teacher-student network to distill knowledge for visual relationship detection. 
Zellers \etal~\cite{zellers2018motif} propose a powerful baseline that infers the relation between two objects by simply taking the most frequent relation between the object pair in training set.
% which outperform both VRD~\cite{lu2016visual} and IMP~\cite{xu2017IMP}. 
By incorporating the frequency baseline, MotifNet~\cite{zellers2018motif} uses bi-directional LSTMs\cite{hochreiter1997lstm} to decode objects and relations sequentially. Recently, Gu \etal~\cite{gu2019knowledge} propose to extract commonsense knowledge of predicted objects from knowledge bases.
% KERN~\cite{chen2019KERN} proposes to use the co-occurrence probabilities of object pairs to describe the correlation between objects, and use this correlation matrix to pass information on the scene graph to obtain more representative features.
Another line of work try to design more sophisticated message passing methods to fine-tune object and predicate features~\cite{qi2019attentive,li2018factorizable,wang2019exploring}. 
Dai \etal~\cite{dai2017relational} propose to extend IMP~\cite{xu2017IMP} by unrolling the iterative inference of a CRF into a deep feed-forward neural network.  
% LinkNet~\cite{woo2018linknet} proposes a Relation Embedding module, which is essentially a self-attention module, to first refine object features and then use object' features to construct and refine predicate features. 
LinkNet~\cite{woo2018linknet} exploits a new Relation Embedding module based on the self-attention to refine object features, which are then used to construct and refine predicate features.
Graph RCNN~\cite{yang2018graph} performs reasoning over the scene graph via a Relation Proposal Network, and then applies a graph convolutional network~\cite{kipf2016GCN} to refine features.
% Graph R-CNN~\cite{yang2018graph} performs reasoning over the scene graph by first designing a Relation Proposal Network that sets the strength of edges and then applies graph convolutional networks~\cite{kipf2016GCN} to refine features. 
 Wang \etal~\cite{wang2019exploring} propose a memory-based module to model objects and relations separately and then pass information between them in an iterative way.

KERN~\cite{chen2019KERN} tries to solve the class imbalance problem  and uses \emph{mean recall@K} as evaluation metric that can better represent the model's performance. However, KERN~\cite{chen2019KERN} tackles this problem by designing a complex network structure, while our proposed framework does not rely on particular network structure and is general and can be easily applied to any scene graph models that train with a  cross-entropy loss, such as KERN~\cite{chen2019KERN} itself. Experiments show that our framework can improve the performance of KERN~\cite{chen2019KERN} by  around 1\% to 2\%.  

Zhan \etal~\cite{zhan2019undetermined} propose a module that classifies each predicted predicate as either determined or not, which is related  to our proposed \ranker\ network. However, their module is simply a stack of fully-connected layers applied to each prediction individually and does not care about ranking, while ours utilizes information from both image-level and  different relations. More importantly, Zhan \etal~\cite{zhan2019undetermined} do not consider ranking and only use their module in training, while our \ranker\ module is used also during inference to generate the ranks of predicted relation triplets.

% On one hand, we employ a self-attention layer to compare one relation triplet with all the other triplets, which shares the spirit of list-wise ranking approaches. On the other hand, we use a binary cross-entropy to learn the \ranker\ network, which is commonly used in point-wise ranking methods.
% Among the three approaches, our proposed \ranker\ network is most related to the \emph{point-wise} approach, where we predict the individual score of each relation triplet by reasoning over other triplets. 
% However, our work is  very different from \emph{learning to rank}, since we do not have ground-truth score for each relation triplet, and our goal is not to predict the exact order but to rank all annotated relations higher than un-annotated ones.

\textbf{Class Imbalance.} This problem has been widely studied in both image classification~\cite{DBLP:journals/corr/abs-1710-05381,japkowicz2002class} and object detection~\cite{oksuz2019imbalance}. The most basic approaches are over-sampling minor classes and under-sampling major classes,  both of which are prone to overfit. Another common approach is to multiply the loss of each class by its inverse  frequency in the dataset. Sampling hard negatives is also widely studied in object detection to address the imbalance problem~\cite{rota2017loss,shrivastava2016training}. Recently, more advanced loss functions have been proposed to address this problem~\cite{lin2017focal,qian2019dr,chen2019towards}. However, the class imbalance problem in scene graph parsing is rarely studied~\cite{chen2019KERN}.  To the best of our knowledge, we are among the first to study this problem in depth and propose a general attentive solution that significantly mitigates the problem.

\textbf{Contrasting Learning.} The contrasting mechanism has been widely used for various applications, by pushing the scores of positive samples or matched sample pairs higher than those of negative ones \cite{chopra2005learning,kim2019unsupervised,wang2015unsupervised,zhang2019learning}. 
Different from these previous approaches, we contrast for each relation the predicted probabilities of the correct and incorrect classes to suppress the incorrect frequent relations.
In other words, they contrast data samples while we contrast classes, and the goals are different.
While \cite{elsayed2018large,hayat2019gaussian} also contrast classes, our design is different from theirs.
They push down all negative classes to maximize the overall discriminativeness while we deliberately  inhibit only the hardest negative class, which is critical to suppress the incorrect frequent relations.
Besides, the contrasting loss in \cite{elsayed2018large} includes a very complex normalization term while our CCE loss is much simpler and computationally efficient.
\cite{hayat2019gaussian} uses Gaussian as the distance metric while we use cross-entropy. 
Zhang \etal~\cite{zhang2019graphical} propose a graphical contrastive loss that discriminate positive \texttt{<subject, object>} pairs from negative ones, which is different from us since our CCE loss contrast the classes of predicted relations.
While all these work focus on one task, to the best of our knowledge, we are the first to apply a contrasting max-margin based loss together with a ranking objective in scene graph parsing.

\textbf{Learning to Rank.} Our work is  also related to the field of \emph{learning to rank}~\cite{liu2009learning}, which aims to predict the ranks of a set of samples in a  non-decreasing order. There are three popular approaches to address this problem, \ie, \emph{point-wise}~\cite{sculley2010combined}, \emph{pair-wise}~\cite{ibrahim2017rank,moon2010intervalrank,sculley2010combined} and \emph{list-wise}~\cite{wang2019list,moon2010intervalrank}. 
%The %Scorer innovatively bridges the point-wise and list-wise ranking approaches to take advantages of both.
%Specifically, it exploits a self-attention layer to compare one relation triplet with all the others in a list-wise fashion, from which learning to rank is achieved via a point-wise cross-entropy loss.
%To our knowledge, this unification is unique and novel. 
Different from prior work, our proposed Scorer network innovatively bridges the point-wise and list-wise ranking approaches by exploiting a self-attention module to compare one relation triplet with all the others in a list-wise fashion, from which learning to rank is achieved via a point-wise cross-entropy loss.

\section{Proposed Approach}
% In this section, we first provide an overview of the scene graph parsing problem, introduce our proposed framework, and then we explain the details of each component in our framework.
% In this section, we first introduce the scene graph parsing problem (Section~\ref{sec:problem}). Section~\ref{sec:framework} provides an overview of the proposed framework. Finally, we explain each component of our approach in details (Section~\ref{sec:cce} and Section~\ref{sec:scorer}).
\subsection{Problem Definition}
\label{sec:problem}
The task of scene graph parsing~\cite{xu2017IMP,zellers2018motif} requires a model to take an image as input and then output an ordered set of relation triplets $\mathcal{T} = \{\mathcal{T}_i| i=1,...,M\}$, where $\mathcal{T}_i=\texttt{<subject, predicate, object>}$ and $M$ is the number of relation triplets in the given image, each \texttt{subject/object} is from a set of object classes $\mathcal{C}$, and each \texttt{predicate} is from a set of predicate classes $\mathcal{R}$. The triplets in set $\mathcal{T}$ are ranked according to their confidence predicted by the model.
% from high to low according to the model's confidence in each triplet's significance in the given image. 

% As current scene graph models are based on pretrained object detectors, the objects' features and bounding boxes are given by models such as Faster R-CNN~\cite{ren2015faster}. 
Current scene graph models rely on pretrained object detectors, \eg, Faster R-CNN~\cite{ren2015faster}, to obtain the bounding boxes of objects and their features.
Given $N$ object proposals from the object detector, the scene graph model predicts their object labels. For each pair of predicted objects, the model generates a relation feature for this \texttt{<subject, object>} pair, which is then classified among the $|\mathcal{R}|$ predicate classes.  Note that there are $N(N-1)$ relation features in total, excluding self-relations. The classification of objects and predicates are trained using separate cross-entropy loss functions in most existing scene graph models \cite{xu2017IMP,zellers2018motif,woo2018linknet,chen2019KERN}. In practice, to handle un-annotated relations, an \texttt{unknown} class is introduced so that classification is performed among $|\mathcal{R}|+1$ classes. During inference, the predicate's label is found by ignoring the Softmax score of the \texttt{unknown} class and taking the label with the largest probability among the rest $|\mathcal{R}|$ classes.

After obtaining the predicted triplets $\mathcal{T}$ and a given set of ground-truth triplets $\mathcal{T}^*$, the result is evaluated by calculating \emph{macro-averaged recall@K}, also known as $mR@K$~\cite{chen2019KERN}:
\begin{align}
    mR@K = \frac{1}{|\mathcal{R}|} \sum_{r\in \mathcal{R}} \frac{\textrm{\# of correctly predicted $r$ in } \mathcal{T}[:K]}{\textrm{\# of $r$ in $\mathcal{T}^*$}},
    \label{eq:recall}
\end{align}
where $\mathcal{T}[:K]$ represents the top-$K$ triplets in the predicted $\mathcal{T}$. This metric is calculated per test image, and then the average $mR@K$ is used as the metric.

\subsection{Overview of Framework}

\label{sec:framework}

Previous scene graph parsing models simply treat the problem as a combination of object and predicate classification problems. However, the evaluation metric, \ie, (mean) recall@K, also requires ranking the detected and classified relation triplets.
To fill this gap, previous models~\cite{xu2017IMP,zellers2018motif,woo2018linknet,chen2019KERN} simply use the product of Softmax probabilities of the predicted \texttt{<subject, predicate, object>} classes to rank the relation triplets. This naive approach empirically works well but is unreasonable since the Softmax scores are normalized per relation\footnote{We use the term "relation" to represent "relation triplet" for saving space.} and only represent the confidence of the model on its classification results. We argue that the confidence in classification results does not necessarily imply the significance of relations in ranking, since a relation's significance is more related to the main content of the scene. For example, the relation triplet  \texttt{<bird, on, flower>} is important in an image consisting of only birds and flowers, but the same relation is trivial in an image where the main content is \texttt{<people, playing, football>}. 
These trivial relations are usually unlabeled in the dataset and will be considered as \emph{negatives} in evaluation.
 Thus we propose to formulate the scene graph parsing problem as a combination of both \emph{classification} and \emph{ranking} problems, and design a novel \emph{Contrasting Cross-Entropy (CCE) loss} and a \emph{\ranker}\ network to address each of these two tasks.

\begin{figure*}[t]
\begin{center}
   \includegraphics[width=0.9\linewidth]{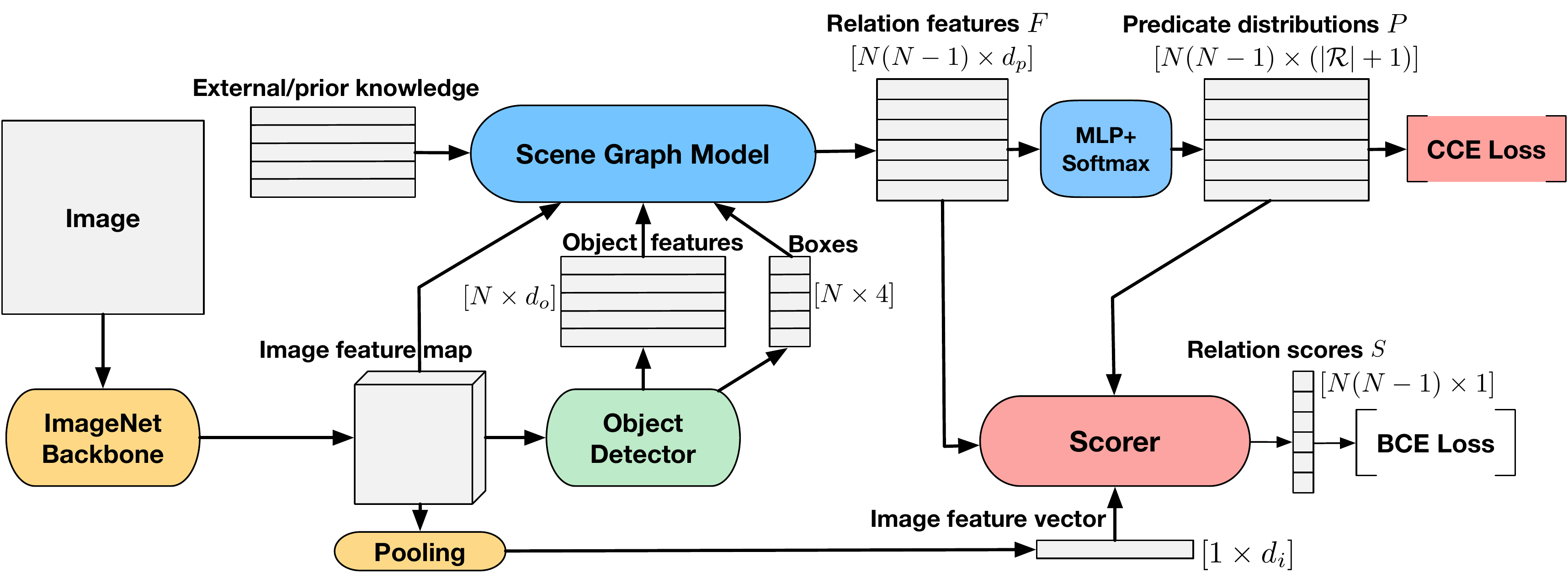}
\end{center}
   \caption{An overview of the proposed framework. Here $N$ is the number of detected objects in the input image, and $(d_o,d_i,d_p)$ are the feature dimensions of (object, image, predicate). The core components are the Contrasting Cross-Entropy (CCE) loss for predicate classification and the \ranker\ network for relation ranking. Since these two components do not require modifying the chosen scene graph model, any existing scene graph model that trains with cross-entropy can be easily plugged into our framework. The external/prior knowledge is specific to the chosen base scene graph model. For example, MotifNet~\cite{zellers2018motif} and many other models~\cite{woo2018linknet,zhang2019graphical} use the probability distribution of predicates given the subjects and objects.}
\label{fig:framework}
\end{figure*}

The overall framework of our proposed method is illustrated in Figure~\ref{fig:framework}. The main pipeline is general among most scene graph models: an image is first fed into a backbone image classifier pretrained on ImageNet~\cite{deng2009imagenet}, and then the output image feature tensor is fed into an object detector pretrained on the Visual Genome dataset~\cite{krishna2017visual_genome} to obtain a set of $N$ object proposals. By using any existing scene graph models as the base scene graph model, we obtain $N(N-1)$ relation features $F \in \mathbb{R}^{N(N-1)\times d_p}$ and their probability distributions $P \in \mathbb{R}^{N(N-1)\times (|\mathcal{R}|+1)}$ among the $|\mathcal{R}|+1$ predicate classes. These predicate distributions are used in our novel CCE loss for predicate classification, while the \ranker\ network takes the relation features, predicate distributions as well as the global image feature to generate the significance scores for each of the $N(N-1)$ relations. The scores are then used to rank the relations during inference.

\subsection{Contrasting Cross Entropy Loss}
\label{sec:cce}
As widely studied, classification models trained with a cross-entropy (CE) loss can be highly biased towards frequent classes and thus perform badly on rare  ones. Methods such as Focal Loss (FL)~\cite{lin2017focal} have been proposed to address this problem in object detection, but the class imbalance problem in scene graph parsing is less explored. In this section, we propose a simple but effective Contrasting Cross-Entropy (CCE) loss that can significantly improve mean recall@K and can be applied to any existing scene graph model using cross-entropy loss. Recall that the cross-entropy (CE) loss for one relation is defined as:
\begin{align}
    \mathcal{L}_{\textrm{CE}}(\mathbf{p}, \mathbf{y}) = -\sum_{i=1}^{|\mathcal{R}|+1} \mathbf{y}_i \log \mathbf{p}_i,
\end{align}
where $\mathbf{p}$ is the probability distribution over the $|\mathcal{R}|+1$ predicate classes including the \texttt{unknown} class, and $\mathbf{y}$ is the one-hot encoded ground-truth label vector.

As minimizing cross-entropy loss is equivalent to maximizing the log probability of the correct class, in order to handle the class imbalance problem, we also force the model to minimize the log probability of the hardest negative class, and design our Contrasting Cross-Entropy (CCE) loss as:
\begin{align}
    \mathcal{L}_{\textrm{CCE}}(\mathbf{p}, \mathbf{y}) =  \textrm{max}(\mathcal{L}_{\textrm{CE}}(\mathbf{p}, {\mathbf{y}}) - \mathcal{L}_{\textrm{CE}}(\mathbf{p}, \hat{\mathbf{y}}) + \alpha,0 ),
\end{align}
where $\hat{\mathbf{y}}$ is the sampled hardest negative label, and $\alpha$ is a hyper-parameter to control the margin. 
Our intuition is that the hardest negative classes usually correspond to frequent relations.
By suppressing the Softmax scores of incorrect frequent classes, this loss function can help the model perform better on rare classes. Here we choose the hardest negative label by first ignoring the \texttt{unknown} class (\textit{i.e.} $\mathbf{p}_1$) and taking the incorrect class with the highest probability, \ie, $c = \argmax_{j > 1, \mathbf{y}_j = 0} \mathbf{p}_j$, and $\hat{\mathbf{y}}=\{\hat{\mathbf{y}}_i|i\in[1,|\mathcal{R}|+1]\}$,where:
\begin{align}
     \hat{\mathbf{y}}_i= 
\begin{cases}
    1, & \text{if } i=c\\
    0,              & \text{otherwise}.
\end{cases}
\end{align}
The hardest negative can be found as we calculate the normalization term for Softmax, so the overhead brought by our CCE loss is ignorable.

% \subsubsection{Adaptive Margin}

% Although the margin $\alpha$ can treated as a hyper-parameter, it may not be optimal to use the same margin for all classes, thus we propose to calculate an adaptive margin based on the pair of positive and negative labels, and we call our CCE loss with adaptive margin as CCE-AM.

% Since each predicate is composed of words, we use pretrained Glove~\cite{pennington2014glove} embeddings to obtain individual word embeddings, and the predicates' embeddings are calculated by the average embeddings of their words. The embedding for the ``unknown'' class is simply defined as the average of embeddings of all other classes.

% Since some predicates are more similar, some are more distinctive,  intuitively, we want to set a small margin for similar classes and larger ones otherwise:
% \begin{align}
%     \alpha(\mathbf{y},\hat{\mathbf{y}}) = \textrm{CosineDistance}(E(\mathbf{y}), E(\hat{\mathbf{y}})),
% \end{align}
% where $E(\cdot)$ is a look-up table containing the embeddings of all predicate classes including the ``unknown'' class.

\subsection{Learning to Score Relations}
% Given a set of detected objects in an image, there is always a relationship between any pair of objects, although this relationship may not be in the given predicates label space or is neglected by annotators. Thus, it is crucial to rank the predicted relations  properly so that annotated relations are ranked higher than un-annotated ones. 

\label{sec:scorer}
In this section, we introduce how we design our \ranker\ network, which produces a score of each predicted relation triplet by utilizing the relation feature and the predicate distribution produced by a base scene graph model.
By exploiting a self-attention module to compare each relation triplet with all the others, it learns to rank the nontrivial relations higher than the trivial ones and thus reduces \emph{false positives}.

%Our \ranker\ means to produce a score for each predicted relation triplet so that the significant and nontrivial relations rank higher than those trivial ones.

While an object may either exist or not in an image, there  is always at least one relationship between any two existing objects, \eg, \texttt{left} or \texttt{right}. When people annotate the SGP dataset, they only label those most important or nontrivial relations but ignore the others. 
As a result, the detection of an existing but trivial relation will be counted as a \emph{false positive} because it is unlabeled in the dataset.
To achieve a high recall for a given number of predictions, we need to not only  classify the relations correctly but also rank the nontrivial relations higher than the trivial ones. This motivates the design of our Scorer which exploits a self-attention module to compare each relation triplet with all the others and learns to rank them.

\begin{figure}[tb]
\begin{center}
   \includegraphics[width=0.7\linewidth]{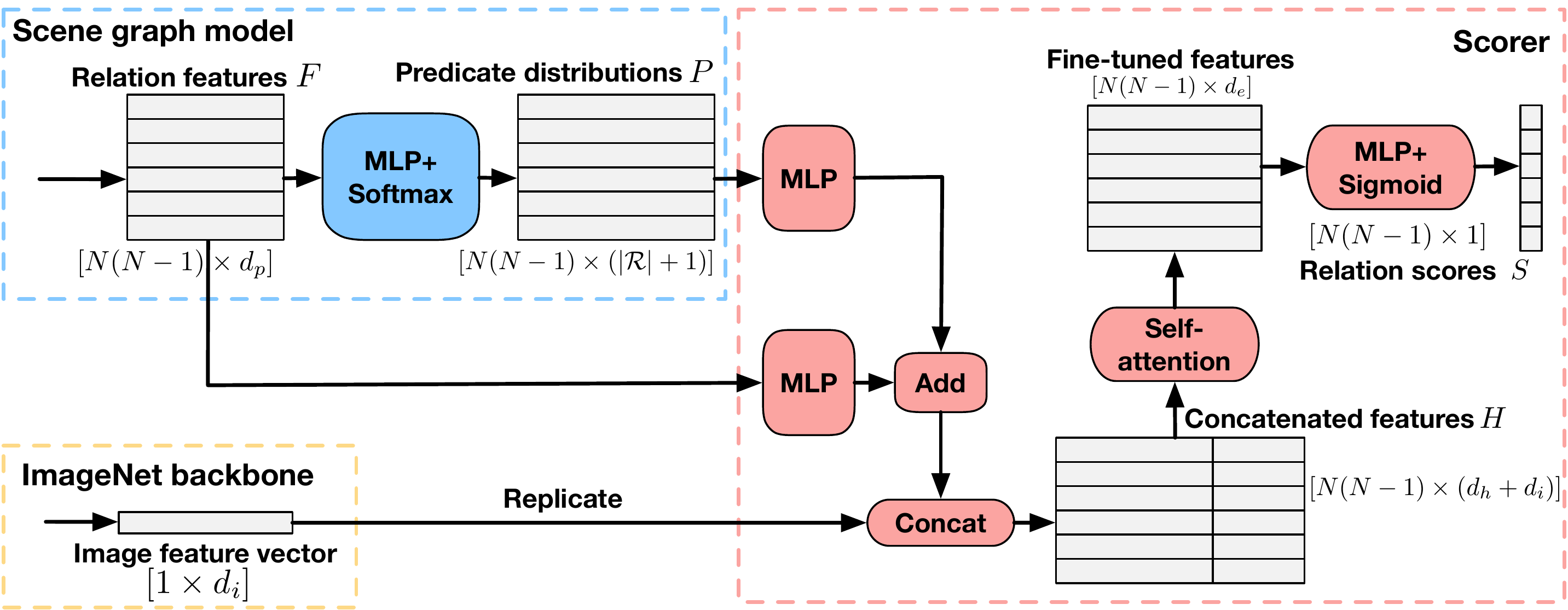}
\end{center}
   \caption{The proposed \ranker\ network that learns to score each predicted relation triplet.}
\label{fig:ranker}
\end{figure}

As illustrated in  Figure~\ref{fig:ranker}, the relation features and predicate probability distributions from the base scene graph model are fed into two different multi-layer perceptron (MLP) networks, where the  two MLPs serve as non-linear transformation functions. The output of two MLPs are added along the feature dimension, and then the input image feature extracted by the ImageNet backbone is concatenated to each relation feature to let the relation feature maintain global information of the image. 

As the significance of a relation is also dependent on the existence of other relations, it is necessary to compare all detected relations before assigning their significance scores. In order to perform such reasoning, the combined relation features are fed into a dot-product self-attention (SA) module~\cite{vaswani2017attention} to pass information among different relations:
\begin{align}
\label{eq:attn}
    \textrm{SA}(H) = \textrm{Softmax}(f_{\textrm{query}}(H)f_{\textrm{key}}(H)^\intercal)f_{\textrm{value}}(H),
\end{align}
where $H \in \mathbb{R}^{N(N-1)\times (d_h+d_i)}$ is a matrix containing relation features concatenated with image features, while $f_{\textrm{query}}(\cdot)$, $f_{\textrm{key}}(\cdot)$ and $f_{\textrm{value}}(\cdot)$ are non-linear transformation functions applied to $H$. Then a final MLP with Sigmoid as its last activation function is applied to obtain the scores $S \in \mathbb{R}^{N(N-1)\times 1}$ which contains the significance score $s$ for each predicted relation triplet.

Although the goal of \ranker\ is to predict the significance scores of relations, we do not have ground-truth scores. Remember that we only need to rank the annotated relations higher than the un-annotated ones (whose predicate labels are given as \texttt{unknown}), 
% and that it makes no difference in $mR@K$ if we rank an annotated relation higher than another annotated one, 
so we treat all annotated relations equally and assign 1 as their significance scores, while all un-annotated relations are assigned with 0 scores. In this case, the \ranker\ module can be trained with a binary cross-entropy (BCE) loss for each predicted relation triplet:
\begin{align}
    \mathcal{L}_{\textrm{BCE}}(s, q) = -q\log(s) - (1-q)\log(1-s),
\end{align}
where $s$ is an output score of \ranker\ and $q=0$ indicates the ground-truth label of this predicate belongs to the \texttt{unknown} class, otherwise $q=1$. We have also tried other loss functions for regression, such as mean-square error and hinge loss, but we found that BCE works best.

During inference, the output scores of the \ranker\ network are used to determine the ranks of predicted relation triplets (the higher the score, the higher the rank), which are then used for calculating $mR@K$.

% Although Zhan~\etal~\cite{zhan2019undetermined} also explore a similar idea of adding an extra module to classify whether each relation is ``known'' or ``unknown'', their module simply applies FCNs to each relation feature individually, thus ignoring the global image information and does not allow reasoning among different relations. Additionally, Zhan~\etal~\cite{zhan2019undetermined} only use their designed module for training, while we also use our \ranker\ network during inference to generate the scores for ranking detected relation triplets.

The whole framework is trained by combining $\mathcal{L}_{\textrm{CCE}}$ and $\mathcal{L}_{\textrm{BCE}}$ and other loss functions $\mathcal{L}_{\textrm{other}}$ specific to the chosen base scene graph model:
\begin{align}
\label{eq:loss}
    \mathcal{L} = \lambda_1 \mathcal{L}_{\textrm{CCE}} + \lambda_2  \mathcal{L}_{\textrm{BCE}} + \mathcal{L}_{\textrm{other}},
\end{align}
where $\lambda_1$ and $\lambda_2$ are two hyper-parameters to balance the two loss terms.

\section{Experiments}
% In this section, we first introduce the  dataset used, evaluation protocol, as well as baseline methods, and then validate the effectiveness of our proposed framework by  quantitative results and ablation study.

\begin{table*}[!t]
\centering
\caption{Quantitative comparison of different methods. *Note that we directly used the KERN~\cite{chen2019KERN} checkpoint trained on SGCls and test on all three tasks without fine-tuning for SGGen, while the KERN paper~\cite{chen2019KERN} finetunes for SGGen. We do not finetune for SGGen since we treat scene graph parsing as an extension of object detection, thus the bounding box regression should be solved by the object detector.}
% \vspace{-0.8em}
\resizebox{\textwidth}{!}{
\begin{tabular}{ l|c c c|c c c|c c c }
\hline
Task & \multicolumn{3}{c|}{SGGen} & \multicolumn{3}{c|}{SGCls} & \multicolumn{3}{c}{PredCls} \\ \hline 
Macro-averaged Recall@K & K=20 & K=50 & K=100 & K=20 & K=50 & K=100 & K=20 & K=50 & K=100 \\ \hline \hline
MotifNet~\cite{zellers2018motif} & 2.09 & 3.37 & 4.52 & 6.22 & 7.62 & 8.06 & 10.87 & 14.18 & 15.31 \\ \hline
LinkNet~\cite{woo2018linknet} & 2.58 & 4.03 & 5.43 & 6.64 & 8.11 & 8.69 & 12.02 & 15.64 & 16.98 \\ \hline
MotifNet+Focal~\cite{lin2017focal} & 1.68 & 3.16 & 4.50 & 5.79 & 7.56 & 8.41 & 10.13 & 14.11 & 16.12 \\ \hline
LinkNet+Focal~\cite{lin2017focal} & 1.51 & 2.99 & 4.62 & 6.1 & 8.41 & 9.52 & 11.02 & 15.85 & 18.35 \\ \hline
KERN~\cite{chen2019KERN} w/o fine-tuning* & 2.24 & 3.96 & 5.39 & 7.68 & 9.36 & 10.00 & 13.83 & 17.72 & 19.17 \\ \hline
\textbf{KERN+CCE+\ranker} & \textbf{2.95} & \textbf{4.85} & \textbf{6.06} & \textbf{9.39} & \textbf{11.28} & \textbf{11.94} & \textbf{16.61} & \textbf{20.57} & \textbf{22.14} \\ \hline
\textbf{MotifNet+CCE+\ranker} & \textbf{3.06} & \textbf{4.72} & \textbf{6.11} & \textbf{7.89} & \textbf{9.91} & \textbf{10.61} & \textbf{15.03} & \textbf{19.62} & \textbf{21.49} \\ \hline
\textbf{LinkNet+CCE+\ranker} & \textbf{3.14} & \textbf{4.96} & \textbf{6.23} & \textbf{9.32} & \textbf{11.53} & \textbf{12.48} & \textbf{17.53} & \textbf{22.23} & \textbf{24.22} \\ \hline \hline
KERN~\cite{chen2019KERN} paper (fine-tuned) & N/A & 6.4 & 7.3 & N/A & 9.4 & 10.0 & N/A & 17.7 & 19.2 \\ \hline
\end{tabular}}
\label{tab:results}
% \vspace{-1.5em}
\end{table*}

\subsection{Experiment Settings}

\textbf{Dataset.} Visual Genome~\cite{krishna2017visual_genome} (VG) is the largest scene graph dataset containing 108,077 images with an average of 38 objects and 22 relation triplets per image. As there exist different splits for the VG dataset, we adopt the most widely used one~\cite{xu2017IMP}, which contains 75,651 images for training (including 5,000 for validation) and 32,422 images for testing. There are numerous objects and predicates in the original VG dataset, but many of them have very low frequencies and quality. We follow Xu~\etal~\cite{xu2017IMP} and only use the most frequent 150 object classes and 50 most frequent predicate classes in VG.

\textbf{Tasks.}
We adopt the most widely studied tasks~\cite{xu2017IMP,zellers2018motif,woo2018linknet,zhan2019undetermined} for evaluation:
    (1) \textbf{Scene Graph Generation} (SGGen) aims to simultaneously localize and classify objects in an image, and predict the potential predicate between each pair of detected objects. An object is considered to be correctly detected if its intersection over union (IoU) with a ground-truth object is over 0.5. 
    (2) \textbf{Scene Graph Classification} (SGCls) provides the model with a set of bounding boxes of objects, and requires the model to predict the object labels as well as the pairwise relations between the objects.
    (3) \textbf{Predicate Classification} (PredCls) requires the model to predict the potential predicates between each pair of objects, where the object locations and labels are given as groundtruth.

\textbf{Metric.} As discussed in previous sections, the widely used \emph{micro-averaged Recall@K (R@K)}  can be highly biased towards frequent classes and cannot tell whether a model is performing well on both frequent and rare classes. Instead, we use the \emph{macro-averaged Recall@K (mR@K)} (Equation~\ref{eq:recall}), \ie, ``mean recall@K''~\cite{chen2019KERN}, as our evaluation metric, which treats major and minor classes alike. We adopt the widely used protocol for determining whether a predicted relation triplet matches groundtruth as in \cite{xu2017IMP,zellers2018motif}.  
% \vspace{-1em}

\subsection{Baselines} 
% \vspace{-0.7em}

We use three of the current state-of-the-art methods \textbf{MotifNet}~\cite{zellers2018motif}, \textbf{LinkNet}~\cite{woo2018linknet} and \textbf{KERN}~\cite{chen2019KERN} as the base scene graph models in our framework and study how our proposed method can help them improve their performance on rare predicate classes. 
KERN~\cite{chen2019KERN} is a recent approach that also tries to solve the class imbalance problem in scene graph parsing by designing a more complex network structure, and also uses the same evaluation metric as we do. 
 However, the authors of KERN train different models for each of the three tasks, while we only train one model for SGCls and test it on all three tasks, so we just used the model checkpoint trained for SGCls provided by the authors and evaluate it directly on all three tasks. We also apply our framework to KERN to show that our proposed method is general and can further help models that already have specific network structures to handle the class imbalance problem. 
As \textbf{Focal Loss}~\cite{lin2017focal} is a popular method that addresses the class imbalance problem in object detection, we also try applying it to LinkNet~\cite{woo2018linknet} and MotifNet~\cite{zellers2018motif} in the scene graph parsing setting to see if it helps alleviate class imbalance in scene graph parsing.

\subsection{Implementation Details}

We use pretrained VGG16~\cite{simonyan2014vgg} as our ImageNet backbone and Faster R-CNN~\cite{ren2015faster} pretrained on VG~\cite{krishna2017visual_genome} as the object detector.  For MotifNet~\cite{zellers2018motif} and KERN~\cite{chen2019KERN}, we use the official code provided by the authors\footnote{https://github.com/rowanz/neural-motifs}\footnote{https://github.com/HCPLab-SYSU/KERN}, and use the same set of hyper-parameters for the models as they are provided in their code base. Although there is no official implementation for LinkNet~\cite{woo2018linknet}, we found an unofficial one\footnote{https://github.com/jiayan97/linknet-pytorch} that produces results close to those in the original paper. 
For Focal Loss~\cite{lin2017focal}, we set $\gamma=4$ and scale the loss by a weight (10 for LinkNet and 100 for MotifNet). 
We apply our proposed CCE loss and \ranker\ to KERN~\cite{chen2019KERN}, LinkNet~\cite{woo2018linknet} and MotifNet~\cite{zellers2018motif} without modifying their original model architectures, and only take the relation features before and after their final MLPs together with the global-average pooled image feature extracted from VGG16 pretrained on ImageNet~\cite{deng2009imagenet} as input to our \ranker\ module. The two MLPs in our \ranker\ are both implemented as one fully connected layer of 512 units and a ReLU activation.  The non-linear transformation functions $f_{\textrm{query}}(\cdot)$, $f_{\textrm{key}}(\cdot)$ and $f_{\textrm{value}}(\cdot)$ in Equation~\ref{eq:attn} are implemented as a fully connected layer with 256 units, and the last MLP in \ranker\ compresses the 256-dim features to scalars followed by a Sigmoid function. \emph{We only tune $\lambda_1$ and $\lambda_2$ for loss weights in Equation~\ref{eq:loss} and keep other hyper-parameters fixed, while we found that setting $\alpha=0$ generally works well for all models}. 
With only two hyper-parameters to tune, our framework is very light-weighted and can be easily applied to other scene graph models. 
For fair comparison, we train all models on the SGCls tasks, and then directly evaluate on all three tasks without fine-tuning for each task. The random seed is fixed as 42 for all experiments. 
Our code is available online\footnote{https://github.com/stevehuanghe/scene\_graph}.

\begin{figure*}[!t]
\begin{center}
   \includegraphics[width=0.99\linewidth]{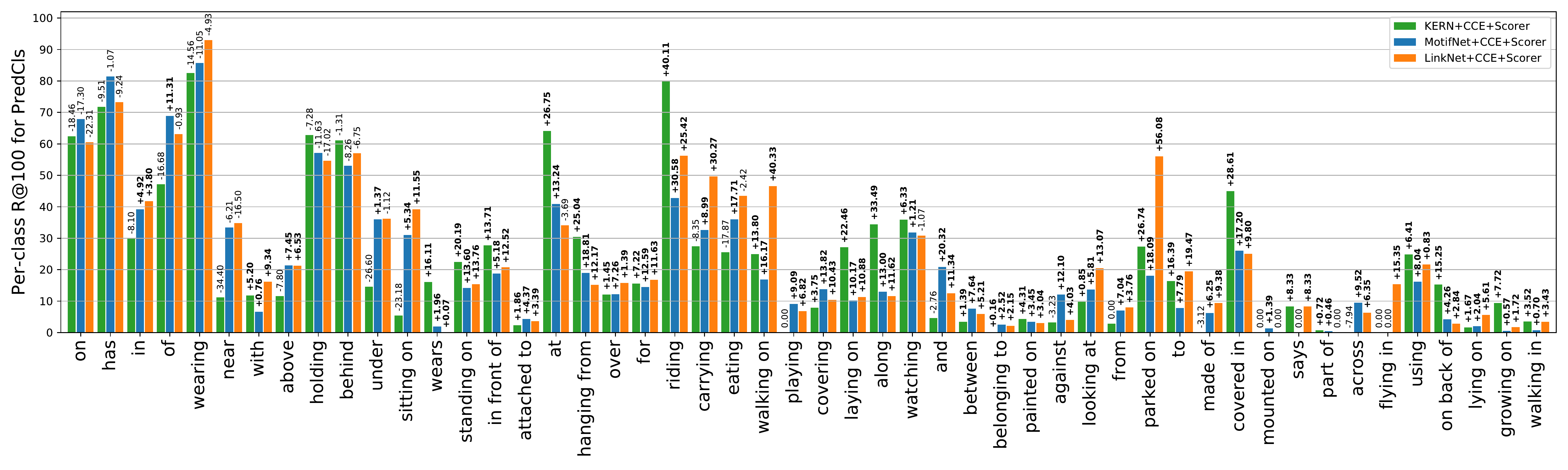}
\end{center}
   \caption{Per-class Recall@100 for KERN~\cite{chen2019KERN}, LinkNet~\cite{woo2018linknet} and MotifNet~\cite{zellers2018motif} with CCE loss and \ranker, evaluated on the PredCls task. The number above each bar indicates the absolute increase ($+$) or decrease ($-$) in each predicate compared with the corresponding base scene graph model trained with cross-entropy. Predicates are sorted according to their frequencies in Figure~\ref{fig:stats}.}
\label{fig:compare-all}
\end{figure*}

\subsection{Results}
We present the quantitative results for our methods together with those of baselines in Table~\ref{tab:results}. As can be seen from the table, although MotifNet~\cite{zellers2018motif} and LinkNet~\cite{woo2018linknet} are among the current state-of-the-art methods that have high \emph{micro-averaged Recall@K} (R@K) of around 67\%, they have relatively low \emph{macro-averaged Recall@K} (mR@K), where their mR@100 are 15.31\% and 16.98\% for the easiest PredCls task. As the tasks become more challenging, their performance for SGCls and SGGen are much worse, which shows that they work badly on detecting rare predicate classes. Among all methods without fine-tuning for SGGen,  our proposed framework with LinkNet~\cite{woo2018linknet} as the base scene graph model achieves the highest mR@K in all three tasks and all $K$s. In the PredCls task, the LinkNet+CCE+\ranker\ method outperforms KERN~\cite{chen2019KERN} by a significant margin of around 4\% to 5\%. For the other two more challenging tasks SGCls and SGGen, LinkNet+CCE+\ranker\ method still outperforms KERN w/o fine-tuning by 2\% and 1\% respectively, which show the effectiveness of our proposed framework. Our method with MotifNet~\cite{zellers2018motif} and KERN~\cite{chen2019KERN} as the base scene graph models also outperform all other methods without our proposed framework, which demonstrates that our framework is not specifically designed for a single model but can help improve the performance of different existing scene graph models trained with cross-entropy loss. It can also be noticed that, our framework with LinkNet~\cite{woo2018linknet} performs better than the one with MotifNet~\cite{zellers2018motif}, which is reasonable as LinkNet~\cite{woo2018linknet} itself outperforms MotifNet~\cite{zellers2018motif} when they are both trained with cross-entropy loss. Our framework with KERN~\cite{chen2019KERN} achieves a performance that lies between the other two base models trained with our framework. 
% which may be due to the more complex architecture of KERN~\cite{chen2019KERN} requiring extra hyper-parameter search and more delicate training techniques.
The reason may be that the architecture of KERN~\cite{chen2019KERN} is more complex and requires extra hyper-parameter search and more delicate training techniques.

As for Focal Loss~\cite{lin2017focal}, we can see that it helps both LinkNet~\cite{woo2018linknet} and MotifNet~\cite{zellers2018motif} improve the results when $K=100$ for all three tasks, but it does not work well with a smaller $K$ such as $K=20$. This result is not surprising, since the Focal Loss does not have an explicit mechanism to rank the predicted relations, and we have discussed in previous sections that the scene graph parsing task requires the model to not only classify predicates correctly but also know how to discriminate more important relations from trivial ones to achieve higher (mean) recall@K. 
% The KERN~\cite{chen2019KERN} model achieves better performance than LinkNet~\cite{woo2018linknet} and MotifNet~\cite{zellers2018motif} with or without the Focal Loss~\cite{lin2017focal}. 
The KERN model fine-tuned for SGGen achieves the highest scores on the hardest task SGGen, while the KERN without fine-tuning performs much worse than the fine-tuned version, which shows that fine-tuning is really important for this task. However, for fair comparison, we only compare the no-fine-tuned version with others.

\begin{table*}[!t]
\centering
\caption{Ablation study for CCE loss and \ranker.}
\resizebox{\textwidth}{!}{
\begin{tabular}{ l|c c c|c c c|c c c }
\hline
Task & \multicolumn{3}{c|}{SGGen} & \multicolumn{3}{c|}{SGCls} & \multicolumn{3}{c}{PredCls} \\ \hline
Macro-averaged Recall@K & K=20 & K=50 & K=all & K=20 & K=50 & K=all & K=20 & K=50 & K=all \\ \hline \hline
MotifNet & 2.09 & 3.37 & 6.81 & 6.22 & 7.62 & 8.31 & 10.87 & 14.18 & 16.19 \\ \hline
MotifNet+CCE & 1.43 & 2.97 & \textbf{7.94} & 4.63 & 6.68 & \textbf{9.17} & 7.87 & 12.91 & \textbf{18.93} \\ \hline
MotifNet+\ranker & \textbf{2.37} & \textbf{3.56} & \textbf{7.23} & \textbf{6.36} & \textbf{7.65} & \textbf{8.34} & \textbf{11.79} & \textbf{14.78} & \textbf{16.71} \\ \hline
\textbf{MotifNet+CCE+\ranker} & \textbf{3.06} & \textbf{4.72} & \textbf{9.21} & \textbf{7.89} & \textbf{9.91} & \textbf{10.93} & \textbf{15.03} & \textbf{19.62} & \textbf{22.57} \\ \hline
LinkNet & 2.58 & 4.03 & 7.87 & 6.64 & 8.11 & 9.02 & 12.02 & 15.64 & 17.86 \\ \hline
LinkNet+CCE & 1.32 & 2.73 & \textbf{8.63} & 5.38 & 8.04 & \textbf{11.28} & 9.07 & 14.30 & \textbf{22.06} \\ \hline
LinkNet+\ranker & \textbf{2.71} & \textbf{4.17} & 7.65 & \textbf{6.79} & \textbf{8.18} & 8.94 & \textbf{12.41} & \textbf{15.62} & 17.77 \\ \hline
\textbf{LinkNet+CCE+\ranker} & \textbf{3.14} & \textbf{4.96} & \textbf{10.32} & \textbf{9.32} & \textbf{11.53} & \textbf{12.94} & \textbf{17.53} & \textbf{22.23} & \textbf{25.30} \\ \hline
\textbf{LinkNet+Ours w/o SA} & \textbf{2.68} & \textbf{4.21} & \textbf{8.99} & \textbf{7.86} & \textbf{10.07} & \textbf{11.57} & \textbf{14.96} & \textbf{19.79} & \textbf{23.15} \\ \hline
\end{tabular}}
\label{tab:ablation}
\end{table*}

We also investigate the changes in each predicate's recall@100 on the PredCls task, as illustrated in Figure~\ref{fig:compare-all}. We can see that our framework can generally improve the performance of base scene graph model in many rare classes starting from \texttt{standing on}, which occupies only 1.1\% of all training samples. More remarkably, our framework with LinkNet as the base model increases the recall for a very rare predicate \texttt{parked on} (0.13\% coverage in the whole dataset) from 0\% to 56.08\%.
Among the top-5 most frequent predicates (\texttt{on}, \texttt{has}, \texttt{in}, \texttt{of}, and \texttt{wearing}) covering 74.5\% of all relations in the dataset, our methods sacrifice some performance in \texttt{on}, \texttt{has} and \texttt{wearing}, which is reasonable and acceptable as we try to focus more on fine-grained predicates instead of these coarse-grained ones. Nonetheless, we also have an increase in a highly frequent class \texttt{in}. 
Among all 50 predicates, our framework helps increase the per-class recall@100 of 29, 37, and 42 classes when trained with KERN~\cite{chen2019KERN}, MotifNet~\cite{zellers2018motif} and LinkNet~\cite{woo2018linknet} respectively, which again demonstrates that our framework can help increase the per-class recall of most predicates, whether they are frequent or not. More importantly, our framework helps improve the per-class recall of 18 of the rarest 20 predicates with MotifNet~\cite{zellers2018motif} and LinkNet~\cite{woo2018linknet}, while the performance of 14 of the rarest 20 classes is also improved for KERN~\cite{chen2019KERN} trained with our framework. Since KERN~\cite{chen2019KERN} is already designed to address the class imbalance problem, it makes sense that our framework does not improve KERN's performance as much as it does with  MotifNet~\cite{zellers2018motif} and LinkNet~\cite{woo2018linknet}.
% When compared with KERN~\cite{chen2019KERN}, from Figure~\ref{fig:compare-all} we can see that, although KERN~\cite{chen2019KERN} performs better than our framework on some large classes such as ``on'', ``has'', ``wearing'' and ``near'', our method has the unique advantage of being able to detect some very rare classes where KERN~\cite{chen2019KERN} has zero recall, such as ``belonging to'', `` painted on'', ``says'', ``flying in'' and ``lying on''. Additionally, our framework also outperforms KERN~\cite{chen2019KERN} in many rare classes by a large margin, such as ``sitting on'', ``riding'', ``carrying'' and ``parked on''.

\subsection{Ablation Study}
% In this section, we study the effect of applying only either CCE loss or \ranker\ to the base models ~\cite{woo2018linknet,zellers2018motif}. 

\textbf{CCE loss} aims to improve the model's ability to classify rare classes, but it is not designed for ranking the predicted relations. In order to study the model's classification ability without being affected by the ranks of predictions, we study the mR@all for three tasks, as shown in Table~\ref{tab:ablation}. As we can see, our CCE loss increases the mR@all for three tasks with two different base scene graph models~\cite{woo2018linknet,zellers2018motif}. The CCE loss works well when K is large, but not when K is small like 20 and 50, since a smaller K will cause the metric mR@K to be more limited by the ranking performance of the model. As CCE is only designed for classification, it is not surprising that it does not work well when the metric is more dependent on ranking performance.

\begin{figure}[!t]
\centering
\subfigure[]{
\begin{minipage}[l]{0.35\linewidth}
\centering
\includegraphics[width=1\textwidth]{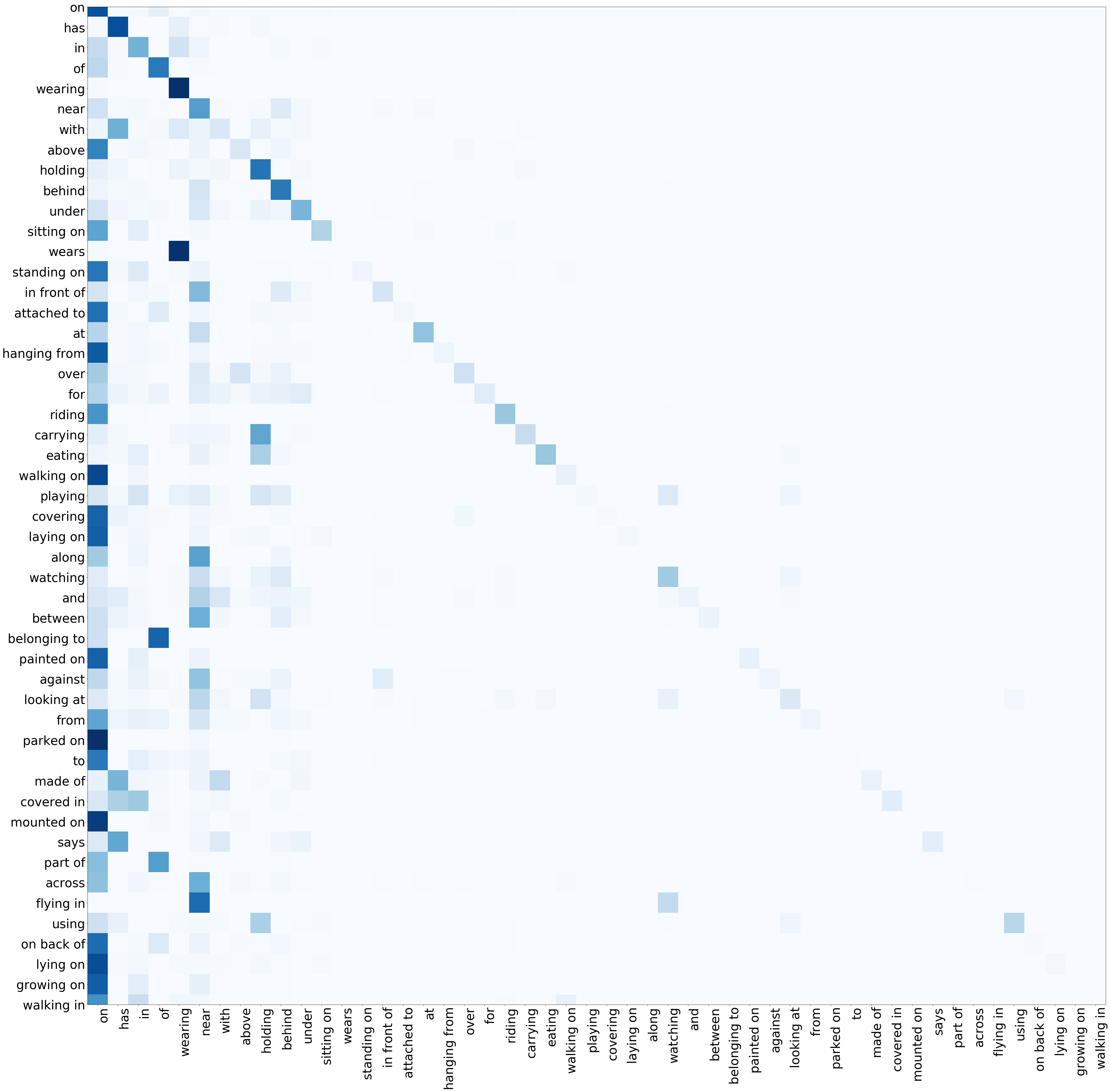}
\end{minipage}
\label{fig:linknet-ce}
}
\subfigure[]{
\begin{minipage}[l]{0.35\linewidth}
\centering
\includegraphics[width=1\textwidth]{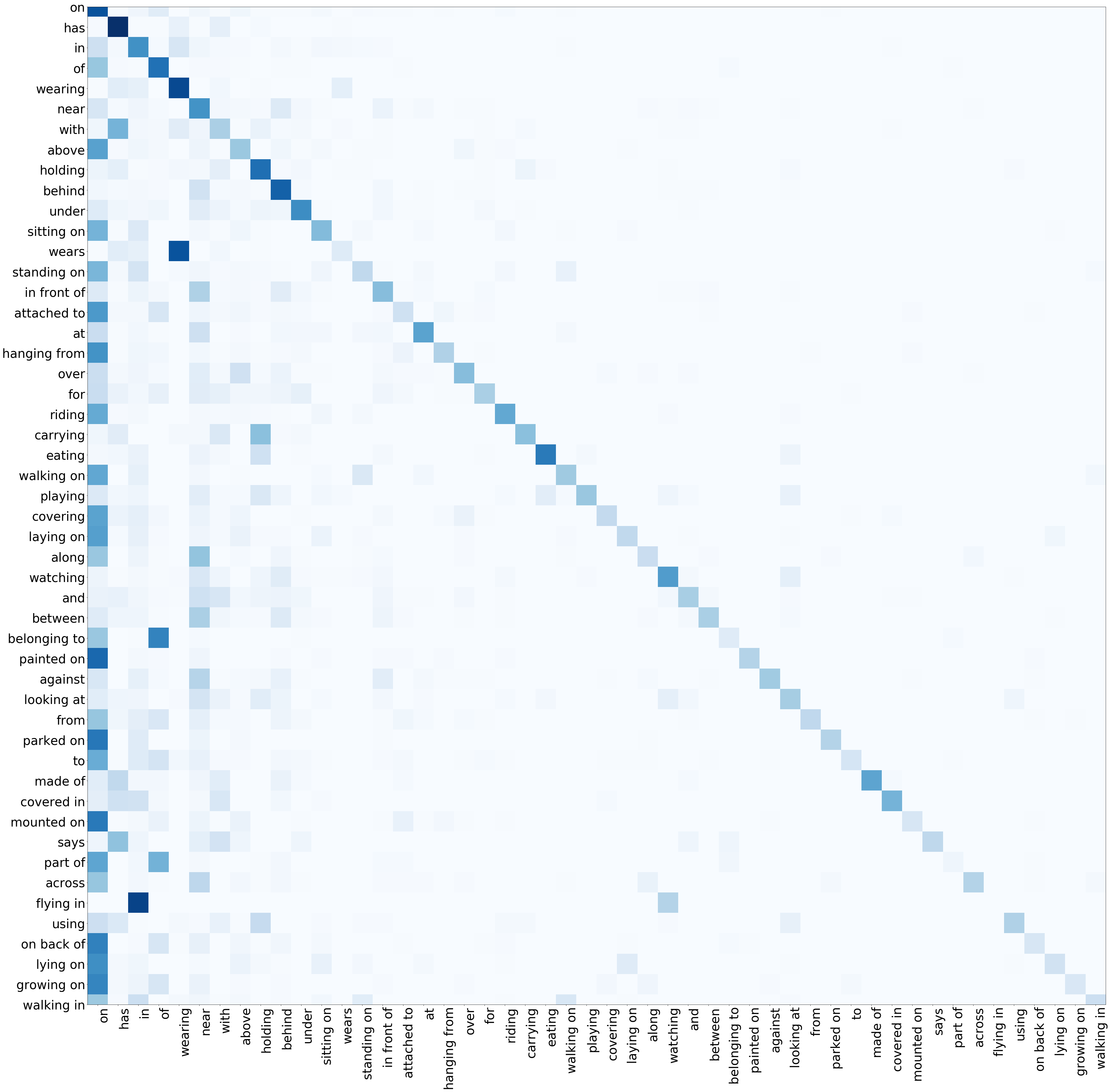}
\end{minipage}
\label{fig:linknet-cce}
}
\caption{Confusion matrix evaluated on VG's training set: (a) LinkNet~\cite{woo2018linknet} with cross-entropy (CE) loss and (b) LinkNet with our proposed contrasting cross-entropy (CCE) loss. The darker the color on diagonal line the better the model. The x-axis shows predicted labels, while the y-axis indicates true labels. Predicates are sorted according to their frequencies in Figure~\ref{fig:stats}.}
\label{fig:cfm}
\end{figure}

We also plot the confusion matrices calculated on the training set to see how well trained models can fit the training data. As shown in Figure~\ref{fig:linknet-ce}, LinkNet~\cite{woo2018linknet} trained with cross-entropy loss is highly biased towards the largest class ``on'', and many points on the diagonal line has shallow color which shows that the model has poor performance in classifying these classes correctly. On the other hand, Figure~\ref{fig:linknet-cce} shows that LinkNet trained with CCE loss achieves better performance on many classes, as the color on diagonal line is much darker than the left one, which shows that the CCE loss can significantly alleviate the class imbalance problem in scene graph parsing.

The \textbf{\ranker}, on the other hand, aims to improve the model's performance in ranking its output relations to achieve higher mR@K when K is small. As shown in Table~\ref{tab:ablation}, our \ranker\ can help improve mR@20 and mR@50 in all cases. For K=all, MotifNet+\ranker\ has small improvements over MotifNet itself, while LinkNet+\ranker\ has a small decrease of average 0.13\% compared with LinkNet~\cite{woo2018linknet} without \ranker, which is reasonable since mR@all is more dependent on the model's classification ability than ranking. Although the base models trained with \ranker\ alone does not have significant improvement over the ones without, it is worth noting that, together with our CCE loss, the whole proposed framework can significantly improve the mean recall@K of both LinkNet~\cite{lin2017focal} and MotifNet~\cite{zellers2018motif} in all three tasks, as shown in Table~\ref{tab:ablation}.

\textbf{Self-attention}. In the last row of Table~\ref{tab:ablation}, we show the results of LinkNet~\cite{woo2018linknet} with our CCE+\ranker~framework but without self-attention, where we simply replace the self-attention module by a fully-connected network applied on each relation triplet individually (pure point-wise ranking). From the results we can see that although the proposed framework without self-attention still outperforms the  LinkNet~\cite{woo2018linknet} baseline, its performance is worse than the one with self-attention (point-wise and list-wise combined ranking). This observation proves that it is important to compare each predicted relation with others by passing information among them, and justifies our design of the \ranker~module that unifies both list-wise and point-wise ranking methods in a novel way.

\textbf{Computational complexity}. The hardest negative class can be found as we calculate the normalization term for Softmax, so the overhead brought by our CCE loss is ignorable. 
 In our implementation, the \ranker~network adds about 3M parameters to MotifNet~\cite{zellers2018motif} (originally 253M), 15M to KERN~\cite{chen2019KERN} (originally 281M), and 4M to LinkNet~\cite{woo2018linknet} (originally 292M). 
 For MotifNet and KERN, our method adds about 6GFlops to their original 218GFlops and 468GFlops, while for LinkNet our method add around 46GFlops operations to its original 326GFlops. The difference is due to the various feature dimensions and the number of relation candidates that the scene graph models produce.

\section{Conclusion}
In this paper, we show that although current scene graph models achieve high \emph{micro-averaged recall@K} (R@K), they suffer from not being able to detect rare classes, and that \emph{macro-averaged recall@K} (mR@K) is a more suitable metric as it treats both frequent and rare classes the same. To tackle the class imbalance problem, we first formulate the task as a combination of classification and ranking, and then we design a framework which consists of a Contrasting Cross-Entropy (CCE) loss and a \ranker\ module respectively for these two sub-tasks. Our proposed framework is general and can be applied to any existing scene graph models trained with cross-entropy. Extensive experiments demonstrate the effectiveness of our framework with KERN~\cite{chen2019KERN},  LinkNet~\cite{woo2018linknet} and MotifNet~\cite{zellers2018motif}. 
% We also conduct ablation study which shows that these two proposed components both serve their intended purposes when used alone. 
% Although our proposed framework can significantly improve recall performance in rare classes, the class imbalance problem in scene graph parsing is still an open problem. 

\section*{Acknowledgement}
This work is supported in part by NSF under grants III-1763325, III-1909323, and SaTC-1930941.

%here would be your acknowledgement (if any) in the final accepted paper

%===========================================================
\bibliographystyle{splncs}
\bibliography{sgbib}

%this would normally be the end of your paper, but you may also have an appendix
%within the given limit of number of pages
\end{document}